\documentclass[10pt,twocolumn,letterpaper]{article}

\usepackage{silence}
\usepackage{iccv}
\usepackage{times}
\usepackage{epsfig}
\usepackage{graphicx}
\usepackage{amsmath}
\usepackage{amssymb}

\newcommand{\DWC}[1]{DWC}

\newcommand{\softP}{\mathcal{P}}

\usepackage{pdfpages}
\usepackage{blindtext}
\usepackage{caption}
\usepackage{textcomp}
\usepackage{gensymb}
\usepackage{stackengine}
\usepackage{graphicx}
\usepackage{capt-of}
\usepackage{booktabs}
\usepackage{varwidth}
\usepackage{mathtools}
\usepackage{floatrow}
\usepackage{float}
\usepackage{esvect}
\usepackage{nicematrix,booktabs}

\newfloatcommand{capbtabbox}{table}[][\FBwidth]
\usepackage{array}
\usepackage{nccmath}
\usepackage[export]{adjustbox}
\usepackage{booktabs,rotating,bigstrut}
\usepackage{bbm}
\usepackage{bm}
\newsavebox\tmpbox
\usepackage{scalerel,amssymb}
\usepackage{subcaption}
\usepackage{cuted}
\usepackage[font=small,labelfont=bf]{caption}
\usepackage{float}
\usepackage{floatrow}
\usepackage{afterpage}
\usepackage{everypage}
\usepackage{arydshln}

\newcommand{\RNum}[1]{\uppercase\expandafter{\romannumeral #1\relax}}
\newcommand\widehatmine[1]{\hstretch{2}{\hat{\hstretch{.5}{#1}}}}

\DeclarePairedDelimiterX{\norm}[1]{\lVert}{\rVert}{#1}

\DeclareMathOperator*{\argmax}{\arg\!\max}
\DeclareSymbolFont{symbolsC}{U}{pxsyc}{m}{n}
\DeclareMathSymbol{\coloneqq}{\mathrel}{symbolsC}{"42}

\usepackage{makecell}

\DeclareCaptionLabelFormat{andtable}{#1~#2  \&  \tablename~\thetable}

\newcommand{\captionaboveof}[3][]{%
    \vskip-\abovecaptionskip
    \vskip+\belowcaptionskip
    \def\@captype{#2}%
    \ifx\@nnil#1\@nnil
        \caption{#3}%
    \else
        \caption[#1]{#3}%
    \fi
    \vskip+\abovecaptionskip
    \vskip-\belowcaptionskip
}
\makeatother

\makeatletter
\newif\ifdelaymatch

\newcommand{\delayfloat}[4]{
  \ifnum\value{page}<#2\relax
    \afterpage{\delayfloat{#1}{#2}{#3}{#4}}%
  \else
    \delaymatchfalse
    \ifcase#3\relax\or
      \if@firstcolumn \delaymatchtrue \fi
    \or
      \if@firstcolumn\else \delaymatchtrue \fi
    \fi
    \ifdelaymatch
      \begin{#1}[t]
        \box#4
      \end{#1}
    \else
      \afterpage{\delayfloat{#1}{#2}{#3}{#4}}%
    \fi
  \fi}

\newif\ifdelaymatchtop

\newcommand{\delaytop}[2]{
  \ifnum\value{page}<#1\relax
    \afterpage{\delaytop{#1}{#2}}%
  \else
    \twocolumn[\box#2\par\vskip\dbltextfloatsep]%
  \fi}

\newenvironment{delayedtop*}[3]{
  \def\delayedtop@box{#3}
  \def\delayedtop@args{{#2}{#3}}%
  \begin{lrbox}{#3}\begin{minipage}{\textwidth}%
    \def\@captype{#1}%
}{
  \end{minipage}\end{lrbox}%
  \global\setbox\delayedtop@box=\copy\delayedtop@box
  \expandafter\delaytop\delayedtop@args
}

\newcommand{\hide}[1]{}

\makeatletter
\renewcommand\paragraph{\@startsection{paragraph}{4}{\z@}%
                                    {1.15ex \@plus0.5ex \@minus.5ex}%
                                    {-1em}%
                                    {\normalfont\normalsize\bfseries}}

\makeatother

\usepackage[pagebackref=true,breaklinks=true,colorlinks,bookmarks=false]{hyperref}

\iccvfinalcopy

\def\iccvPaperID{1579}

\ificcvfinal\fi

\begin{document}

\title{Deep Weighted Consensus (DWC) \\ Dense correspondence confidence maps for 3D shape registration}

\author{Dvir Ginzburg\\
Tel Aviv university\\
{\tt\small dvirginzburg@mail.tau.ac.il}

\and
Dan Raviv\\
Tel Aviv University\\

{\tt\small darav@tauex.tau.ac.il}
}

\maketitle

\begin{abstract}
We present a new paradigm for rigid alignment between point clouds based on learnable weighted consensus which is robust to noise as well as the full spectrum of the rotation group.

Current models, learnable or axiomatic, work well for constrained orientations and limited noise levels, usually by an end-to-end learner or an iterative scheme. However, real-world tasks require us to deal with large rotations as well as outliers and all known models fail to deliver.

Here we present a different direction. We claim that we can align point clouds out of sampled matched points according to confidence level derived from a dense, soft alignment map. The pipeline is differentiable, and converges under large rotations in the full spectrum of SO(3)\footnote{The 3D rotation group in $\mathbb{R}^3$.}, even with high noise levels.  We compared the network to recently presented methods such as DCP, PointNetLK, RPM-Net, PRnet, and axiomatic methods such as ICP and Go-ICP. We report here a fundamental boost in performance.

\end{abstract}
\section{Introduction}

\begin{figure}
\begin{center}
\hspace{-22pt}
\setlength\tabcolsep{1.5pt}
\begin{tabular}{>{\footnotesize}p{1.3cm}cc}

Input        & \includegraphics[scale=0.15,valign=c]{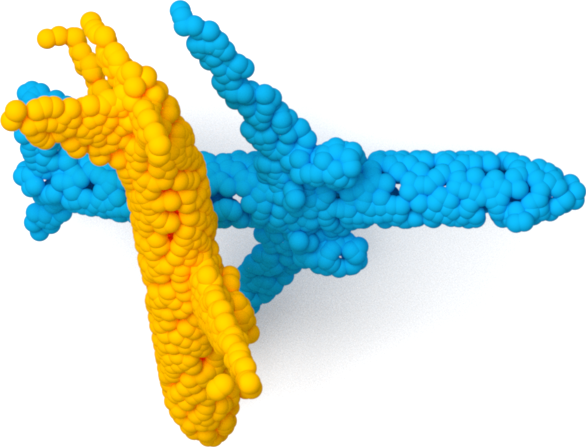}  &  \includegraphics[scale=0.07,valign=c]{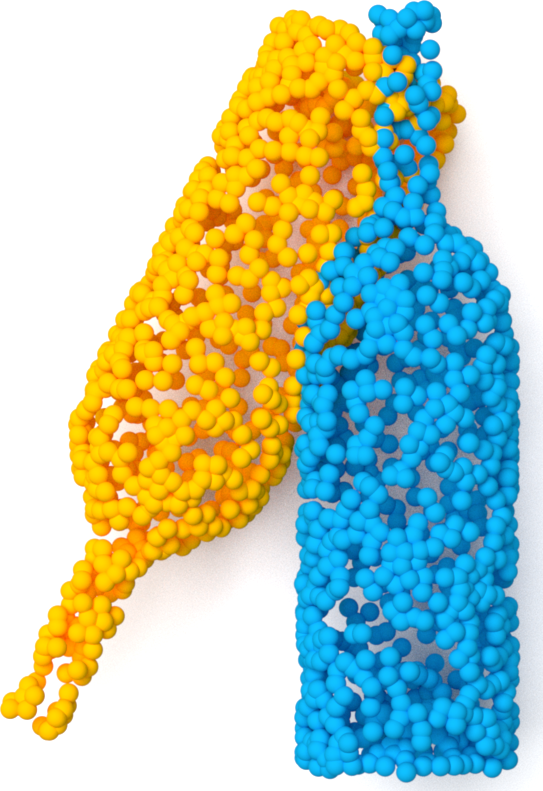}        
\\

ICP   \cite{icp}      &  \includegraphics[scale=0.15,valign=c]{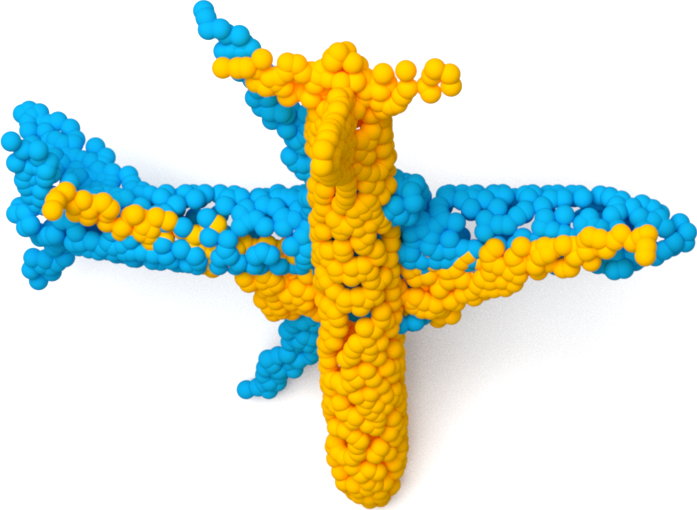}       &  \includegraphics[scale=0.07,valign=c]{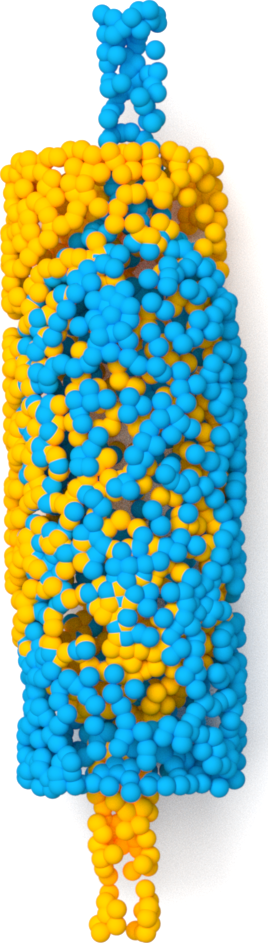}        
\\

DCP  \cite{dcp}      &\includegraphics[scale=0.15,valign=c]{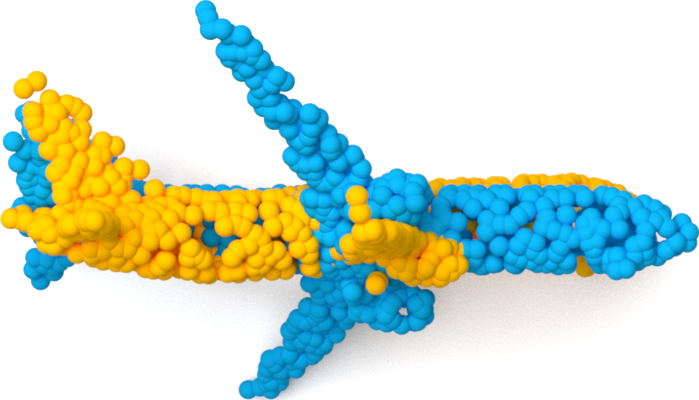}  &  \includegraphics[scale=0.07,valign=c]{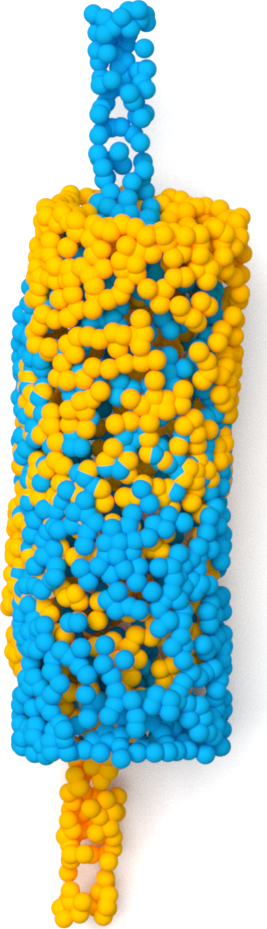}        
\\

PRNet \cite{prnet}        &  \includegraphics[scale=0.15,valign=c]{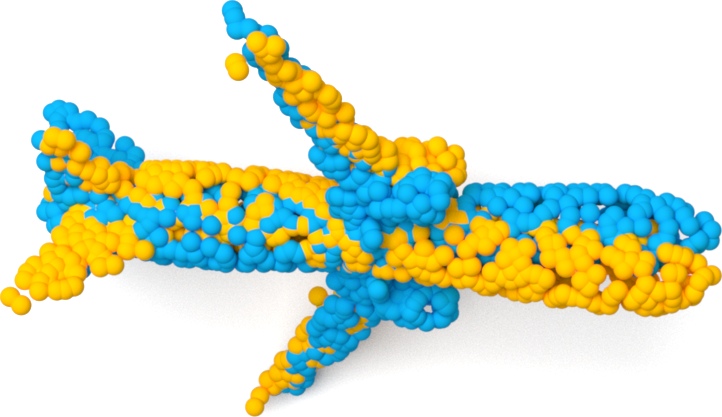}         &  \includegraphics[scale=0.07,valign=c]{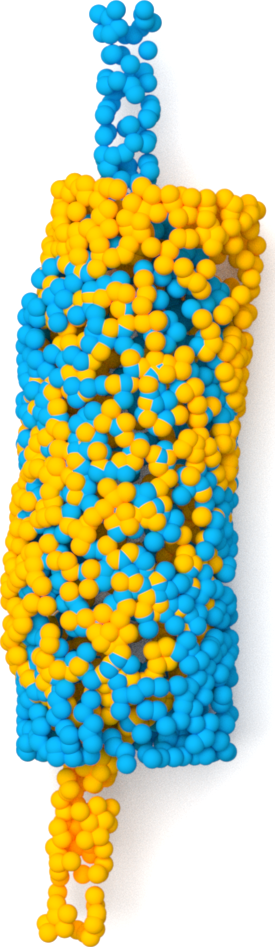}    \\

DWC (ours)         &  \includegraphics[scale=0.15,valign=c]{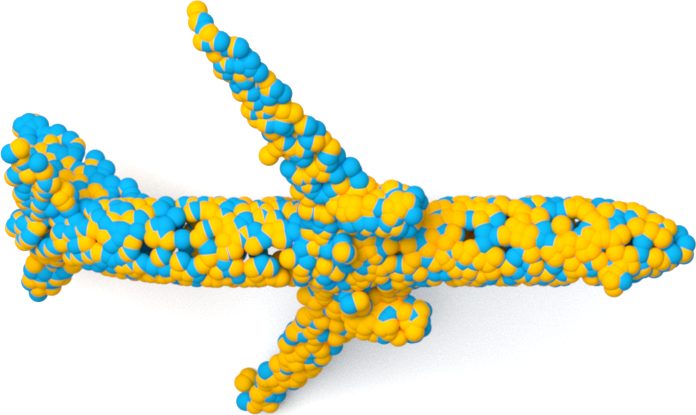}         &  \includegraphics[scale=0.07,valign=c]{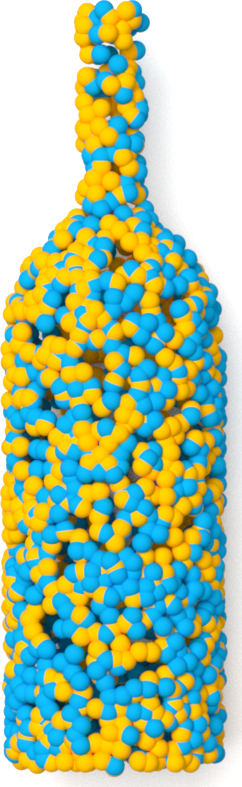}  
\end{tabular}
\end{center}
\caption{Rigid alignment for shapes with intrinsic symmetries and large deformations. DWC is the only method that finds the globally optimal transformation in the full spectrum of SO(3).} \label{fig:Tizer_pairs}

\end{figure}
Alignment between 3D objects is a fundamental problem in computer vision, playing an important role in medical imaging \cite{medical_registration}, autonomous driving \cite{autonomous_registration}, robotics \cite{robotics_registration}, etc. 
At the root of the rigid alignment problem, is the need to find the linear transformation inducing the difference between the objects, where noise, sparsity, partiality, and extreme deformations can distort the input shapes.

The task is even harder in the 3D domain, where data is usually represented by point clouds, thus having irregular topology, varying number of neighbors per point, and unordered data description.
A canonical work in the field is the Iterative Closest Point (ICP) algorithm \cite{icp},  which aligns point clouds in the Euclidean space in iterations until convergence, by matching points according to spatial proximity and then solving a least-squares problem for the global transformation \cite{rtsolve}.
However, ICP is extremely sensitive to noise and initial conditions, and the fact that the problem is non-convex causes ICP to get stuck in sub-optimal local solutions in most cases.
In recent years, many methods \cite{dcp,pointnetlk,prnet,itnet}, presented deep-learning-based improvements to the original idea of ICP. Among the renowned methods is Deep Closest Point (DCP) \cite{dcp}, which replaced the Euclidean nearest point step of ICP by a learnable per-point embedding network, followed by a high-dimensional feature-matching stage.
While improving ICP substantially, all methods mentioned above still use the globally optimal least-squares solution to the transformation problem, thus sensitive to noise, outliers, and errors in the feature-alignment step.

We herein provide a new line of thought for solving the rigid alignment challenge.
We introduce a single feed-forward network both at training and inference by having a voting mechanism on top of a learnable confidence guided sampling step of source points.
Our framework shows superior results in all common test configurations, and even larger improvements in performance when examining rotations on the full spectrum of SO(3)\footnote{rotations larger than $ 60\degree$.}.

In this work, we claim that dense mapping between shapes is not only a \textit{proxy} to the rigid alignment task but provides useful information about the credibility of each point that participate in the process. Here, we optimize for a dense correspondence mapping directly using a new unsupervised contrastive loss in search of alignment and confidence in a fully derivable model.

In our experimentation (Section \ref{subsec:modelnet}) we exemplify the state-of-the-art results our network provides, in the ablation study (Section \ref{subsec:ablation}) and supplementary we demonstrate the significance of each of the three network components, as well as present run-time performance, convergence time, and visual comparisons.

\paragraph{Contributions}
Our contributions include the following:

\begin{itemize}
\itemsep0em 
\item Present a paradigm shift for learning rigid alignment, where the dense correspondence is being optimized directly based on the confidence each point has in its mapping.   
 
\item Report SOTA results on multiple datasets in a substantial number of tests, with comparison to classic and recently published methods. 

\item  DWC is a single feed-forward model during training and testing, converging up to 10 times faster than current state-of-the-art architectures, with a real-time performance at inference.

\item Support the full spectrum of SO(3) without any decrease in performance (RMSE error), unlike previously published methods.
\end{itemize}

\section{Related Work}

\begin{figure*}[ht!]
  \centering
  
  \includegraphics[width=1.\linewidth]{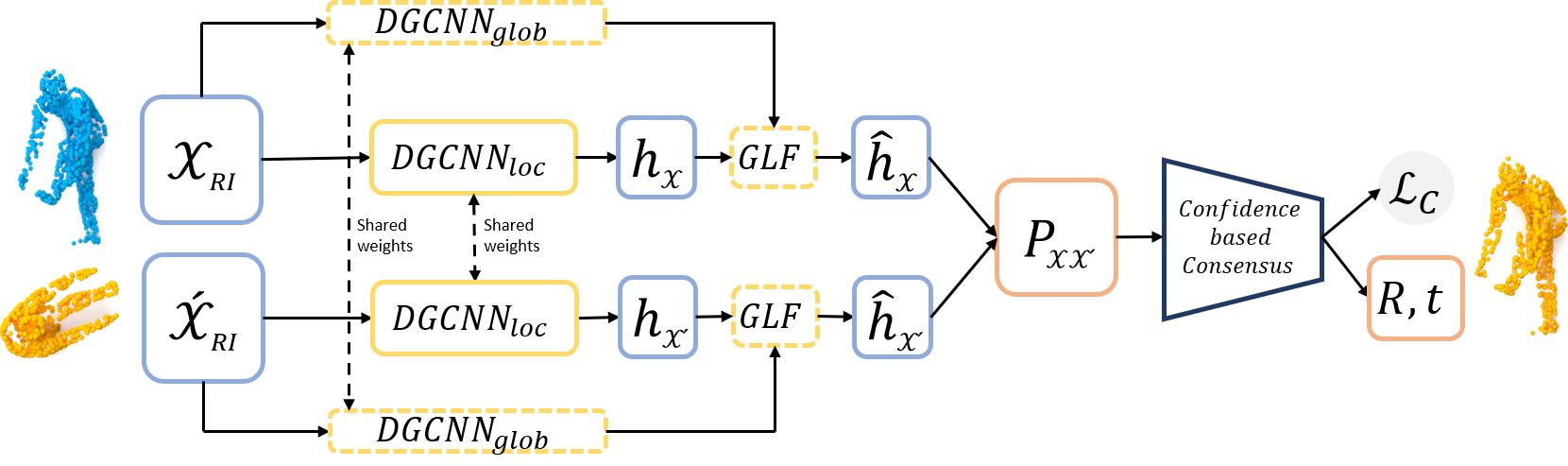}
  \caption{DWC is comprised of three steps: (a)  Embedding linear deformation invariant features ($\mathcal{X}_{RI}$) into a high-dimensional latent space by a global-local fusion network (GLF).  (b) Construct
the soft correspondence map ($\softP_{\mathcal{X}\mathcal{\grave{X}}}$) of the point clouds, defined by
the cosine similarity between the point features, and (c) Confidence based sampling procedure derived from $\softP$. We optimize the soft alignment directly using a weighted contrastive loss $\mathcal{L}_C$ determined by the confidence sampling step.
The transformation parameters are computed only at inference based on the sampled points.
}
  \label{fig:architecture_scheme}
\end{figure*}

\paragraph{Axiomatic point cloud registration}
Axiomatic methods for 3D rigid registration are present for more than 40 years \cite{icp,chen1992object,zhang1994iterative,debevec1996modeling}, with the most renowned of all - the Iterative Closest Point algorithm \cite{icp}. ICP presents an iterative solution to the problem which alternates between two steps, finding correspondences between the two point clouds, and calculating the transformation that aligns the corresponding points followed by applying the transformation. These steps are repeated until convergence but are prone to reach a local-minimum.

ICP variants had tried to solve several problems of the original method such as sensitivity to noise or input sparsity \cite{icpv1,icpv2}, with the drawback of being slower and demanding more computing resources.
The most prominent problem in these solutions is the convergence to local minima and the severe sensitivity to the initial conditions. While many methods have tried to address these issues ~\cite{icpl1,icpl2}, the instability given the input transformation was still present, with running times slower than the original method by up to two orders of magnitude.

\paragraph{Graph neural networks}
The term Graph neural networks (GNNs), coined in \cite{scarselli2008graph}, refers to the family of neural architectures designed for graph-structured data.
\cite{duvenaud2015convolutional} defines the convolution operation on graphs in an analogous way to the definition of CNNs for images, whereas here, unlike grid-structured data, new obstacles arise as varying numbers of neighbors, and irregularity in data order.
Among the class of deep graph neural networks are architectures that work on geometric data, as point clouds and meshes.

PointNet \cite{pointnet} was the first to show how a simple Multi-Layer Perceptron (MLP) per point followed by global pooling can extract meaningful high dimensional features for both classification and segmentation tasks. PointNet++ \cite{pointnetpp} and DGCNN \cite{dgcnn} who came after, exemplified the strength of the concept of message-passing, by aggregating neighbor points features using another learnable function. While PointNet++ uses the Euclidian $k$-nearest neighbors as the graph topology throughout training, DGCNN defines the graph structure dynamically, by the $k$-nearest neighbors in the high-dimensional space per layer. PointCNN \cite{pointcnn} applies a learnable transformation on the point cloud, followed by Euclidean convolution, while \cite{sonet} presents a hierarchical feature extraction for point-clouds, and \cite{kpconv} offers adaptive convolution kernels based on the point neighborhood density and topology.

\paragraph{Learning point cloud registration}
Harvesting the power of graph neural networks, many new methods trying to solve the rigid registration problem were proposed, providing significant improvements over axiomatic techniques.
One such milestone is Deep Closest Point (DCP) \cite{dcp}, which proposed a feature learning scheme followed by a soft alignment of the two point clouds. While DCP suggested many new ideas in the domain, robustness to outliers was still an issue as it took all points into account when evaluating the transformation.
Following DCP, many iterative methods \cite{prnet,pointnetlk,itnet} extended the feature matching idea, where the general scheme was to learn the mapping, apply the inferred transformation on the source point cloud and learn a new alignment map until convergence.
Among these methods is PRNet \cite{prnet}, which tries to solve the outlier sensitivity by choosing a subset of points in the source and target shapes and evaluates the correspondence between the subsets.
While PRNet reduces the chance for outliers, it suffers from accuracy degradation due to the limitation of their sampling. Indeed, PRNet is not on-par with state-of-the-art results in the field.

A significant drawback in all iterative methods is the run-time performance, which linearly depends on the number of refinement steps, making some not suitable for real-time applications.\\
Lately, a new set of methods \cite{deepglobal,robustpointcloud} offered using a fusion of learned features with spatial coordinates throughout training, as initially introduced in PointNet++ \cite{pointnetpp}, to boost the results. While these methods indeed show relative improvements compared to the previously presented works, using spatial coordinates in deeper stages of the network harms the ability to work under considerable transformations, such as when rotations grow higher than 90 degrees, which are common in real-world applications.\\ 
 
\vspace{-5mm}
\section{Deep Weighted Consensus}
\vspace{-2mm}
Deep Weighted Consensus is a self-supervised deep network designed to solve the rigid alignment problem of 3D shapes.

In short, DWC extracts rotation-invariant features of the input point clouds, and builds local and global features for the two shapes. The fusion of these high-dimensional features define the \textit{soft correspondence mapping} between the shapes by computing the cosine-similarity of the features. Next, the network defines a probability distribution based on the certainty of each source point in its map, and sample confident points that partake in the consensus problem. At inference, DWC uses the sampled points to find the optimal $R,t$ (see figure \ref{fig:architecture_scheme} for illustration). 

\subsection{Method}
\paragraph{Rotation invariant descriptors}\label{subsec:rot_inv_feat}
The first step of the DWC pipeline is to extract rotation-invariant ($RI$) descriptors out of the Euclidean coordinates of the input point clouds. These features are the input descriptors for the Feature Extractor Network ($FEN$).
The registration problem in the full range of SO(3) is hard and state-of-the-art methods such as \cite{dgcnn,pointcnn,sonet} fail to generate meaningful representations under such deformations, as proven in \cite{rot1,rot2} and demonstrated in the experimentation shown in section \ref{section:Experimentation}.

One reason for the struggle of networks using the Euclidean coordinates as the inputs for the $FEN$ is the predicted transformation output for symmetric shapes. On symmetric shapes transformed by a $~180\degree$ rotation, such networks will suggest the exact opposite transformation to the correct rotation, as locally, symmetric parts have identical embeddings given the same input features. Illustration to the phenomena can be seen in the plane example in figure \ref{fig:Tizer_pairs}.

Inspired by the local rotation invariant features presented in \cite{rot1}, we present global and local $RI$ features for every point $x$ conditioned by the neighborhood of a source point $m$, which for point clouds is usually the $k$-nearest neighbors.

Our features do not need normal approximations as \cite{rot3}, or geometric median computation as \cite{rot2}, hence are suitable for real-time applications as demonstrated in the timing evaluation provided in the supplementary.\\
Given the Euclidean center of the point cloud $O$, a source point $p$ of neighborhood $\mathcal{N}_p$, and $m$, the Euclidean center of $\mathcal{N}_p$, the rotation invariant descriptors of $x\in \mathcal{N}_p$ are:
\begin{table}[H]
\resizebox{0.8\columnwidth}{!}{%
\begin{tabular}{l|l}
\multicolumn{1}{c|}{Local} & \multicolumn{1}{c}{Global} \\
 $\alpha_{xmp}=\angle(\vv{mx},\vv{mp})$ & $\alpha_{xOp}=\angle(\vv{Ox},\vv{Op})$
 \\
 $\alpha_{xpm}=\angle(\vv{px},\vv{pm})$ & 
 $\alpha_{xOm}=\angle(\vv{Ox},\vv{Om})$
 \\
 $d(x,m)=||x-m||_{2}$     &   $d(x,O)=||x-O||_{2}$     \\
 $d(x,p)=||x-p||_{2}$     &        \\
  
\end{tabular}%
}
\end{table}
\noindent  Where $\vv{aa'}$ is the vector from $a$ to $a'$ in the 3D space and $\angle{(\vv{ab},\vv{ac})}$ is the angle between $\vv{ab}$ and $\vv{ac}$.\\
The aggregated $RI$ feature is:
\begin{align}\label{eq:ri_features}
\begin{split}
    f_{\mathcal{N}_p}(x)=[&\alpha_{xmp},\alpha_{xpm},d(x,m),d(x,p),\\&\alpha_{xOp},\alpha_{xOm},d(x,O)]
\end{split}
\end{align}

We exemplify the significance of using the $RI$ features in the results (Section \ref{subsec:modelnet}) as well as in the ablation study (Section \ref{subsec:ablation}).

\paragraph*{Feature Extraction Network}\label{subsec:feature_gen}
The objective of the feature extraction network is to embed the unaligned input point clouds $\mathcal{X}$ and $\mathcal{Y}$ to a common high dimensional feature space. $FEN$ does this by combining two variants of DGCNN \cite{dgcnn}, namely the classification and segmentation networks, henceforth noted by $DGCNN_{glob},DGCNN_{loc}$ respectively. 
For both networks, we use the representations generated before the last aggregation function, and define them such that for point cloud $\mathcal{X}$ with $N$ points, the output features dimensions are $h_{loc}(\mathcal{X})\in \mathbb{R}^{N\times l_1}$ and 
$h_{glob}(\mathcal{X})\in \mathbb{R}^{l_2}$ where $l_1,l_2$ are network hyper-parameters.

As opposed to the original paper \cite{dgcnn} or subsequent works \cite{dcp,dvir3}, $FEN$ does not use the dynamically changing neighborhood of the graph when applying convolutions in deeper layers of the network, as it causes message passing between distinct parts of the graph, that may have similar characteristics. While this is a good quality for \textit{segmentation} networks, identifying graph-symmetries is crucial for solving the alignment problem. Numerically, we have found that this change improves results substantially (Section \ref{subsec:ablation}).

The final step of $FEN$ is to fuse the global and local features per point. Given $h_{glob}(\mathcal{X})$ and $h_{loc}(\mathcal{X})$, we concatenate the global feature with each local feature resulting in $h_{GL}\in \mathbb{R}^{N\times (l_1+l_2)}$, and propagate $h_{GL}$ through a series of dense non-linear layers, resulting in $\hat{h}_{\mathcal{X}}\in \mathbb{R}^{N\times l_3}$. We mark the Global-Local-Fusion process by GLF.
A detailed layout of $FEN$ is provided in the supplementary.

\paragraph{Soft correspondence mapping}\label{subsec:softcorr}
DWC has the ability to match key points that are similar between $\mathcal{X}$ and $\mathcal{Y}$ with high probability and extract the transformation using these points. Therefore, DWC avoids using outliers in the transformation computation with high probability, in contrast to previous works \cite{dcp,pointnetlk}.
To find these keypoints we define the soft correspondence mapping $\mathcal{P}$, which is given by the cosine similarity of the latent representations of $\hat{h}_{\mathcal{X}}, \hat{h}_{\mathcal{Y}}$: 
\begin{equation}
  \softP_{i,j}=\frac{ \hat{h}_{\mathcal{X}_i} \cdot \hat{h}_{\mathcal{Y}_j}} {|| \hat{h}_{\mathcal{X}_i} ||_2 \cdot ||\hat{h}_{\mathcal{Y}_j}||_2}
\end{equation}
 
$\softP$ is a pseudo\footnote{$\softP$ contains negative values and the rows do not sum to 1.} probability matrix, where $\softP_{ij}$ can be interpreted as the probability that $\mathcal{X}_i$ corresponds to $\mathcal{Y}_j$.\\
Using $\softP$ we define the point-wise mapping between the source and target shapes $\pi: \mathcal{X} \rightarrow \mathcal{Y}$ to be: 
\begin{equation}
    \pi(\mathcal{X}_i)=\argmax_j \softP_{ij}.
\end{equation}
In the theoretical case of perfect samples, given 3 non-collinear points that are correctly matched between the graphs by $\pi(\mathcal{X})$, a closed-form solution to the problem is feasible. In the case of noise and outliers in the mapping, an optimal solution in the least-squares sense was suggested in the Kabsch algorithm \cite{rtsolve}, which solves for $R,t$ given the singular value decomposition of $\pi(\mathcal{X})$.

In the paragraphs that follow, we explain how a sampling procedure based on $\softP$ defines the subset of points that we use as input to the Kabsch algorithm.

\paragraph{Confidence based sampling}
The soft correspondence mapping defines a dense alignment between the shapes. DWC presents a differentiable reduction procedure from $\softP$ to the rigid alignment solution.
Here, unlike previous works that used all points in $\mathcal{X}$ \cite{dcp,pointnetlk}, or chose a fixed subset of points $\{x_i|x_i\in \mathcal{X}\}$ using a $top-k$ heuristic \cite{prnet}, we construct a categorical distribution over the source points, by which we sample points that participate in the solution to the alignment problem.

We define the \textit{confidence metric} $\softP_m$ to be :
\begin{equation}
    \softP_{m_i} = \max_j \softP_{i,j}.
\end{equation}

$\softP_{m} \in \mathbb{R}^{|\mathcal{X}|}$ is a proxy to the confidence level of $\mathcal{X}$ in $\pi(\mathcal{X})$, that is, for point $i$ where DWC has high certainty in the mapping, $\softP_{m_i}$ will be high, while for points where the network is not certain, $\softP_{m_i}$ will be low\footnote{While not common, for $\max_j \softP_{i,j}<0$ we set $\softP_{m_i}=0$.}. 

We normalize $\softP_m$: 
\begin{equation}
   \widehatmine{\softP_m} = \frac{\softP_{m}}{\sum\limits_{i=0}^{|\mathcal{X}|-1} \softP_{m_i}} 
\end{equation}
and define the confidence sampling distribution $S$ over the points in $\mathcal{X}$:

\begin{equation}
\resizebox{0.7\linewidth}{!}{%
\begin{tabular}{cc}
$x_i\in \{i|0\leq i < |\mathcal{X}|\}$ & Support\\[0.2cm]
$s_{X}(x_i)=S(X=x_i)=\widehatmine{\softP_{m_i}}$ & PMF
\end{tabular}%
}
\end{equation}
In what follows we explain how $S$ defines both the registration solution and the optimization objective.

\paragraph{Confidence based consensus}\label{subsec:consensus}
Based on the confidence distribution over the source points $S$,
DWC samples $Q$\footnote{$|Q|$ is a hyper-parameter and was set to $\frac{|\mathcal{X}|}{10}$ in the experiments.} source points. These points are used both for the optimization of the dense alignment, and for the solution of the optimal $R,t$ parameters using the SVD decomposition.
Using the described formulation we allow the network to learn which are the most significant points for the matching and factor them accordingly.

At inference, we sample $k$ "experiment" groups from $Q$ uniformly, each with $r=|Q|/k$ points, where $r$ is the number of points fed to the alignment solver. By sampling uniformly from $Q$, each "experiment" is distributed according to $S$, the confidence based distribution.
We set the chamfer distance \cite{chamfer} between the target shape and the source shape after applying the inferred transformation as the evaluation metric. The extracted $R,t$ parameters for the experiment with the lowest chamfer distance are chosen as the transformation parameters.
Although our inference scheme can be seen as a weighted variant of RANSAC \cite{ransac}, a major difference is the fact that unlike previous methods that used RANSAC, our consensus voting is confident based, and takes part of the learning pipeline itself, and not only during test-time. The incorporation of our weighted consensus problem into the network leads to faster convergence and improved results, as demonstrated in the experiments (Section \ref{section:Experimentation}).

The fact that DWC performs $k$ separate experiments for $R,t$ is not only beneficial to achieving the highest score among all state-of-the-art methods but also provides a leap in performance due parallel computing utilization; solving $k$ $SV\!D$ problems with $m$ equations is up to two orders of magnitude faster than solving one $SV\!D$ problem with $kl$ equations, as done by \cite{prnet,dcp,pointnetlk}.

\subsection{Loss}\label{subsec:losses}

DWC optimizes the feature correlation directly by presenting three constraints on the soft correspondence matrix.
During our self-supervised training, each source shape is being augmented by rotation, translation, and noise. As the same points are "chosen" for both source $\mathcal{X}$ and target $\mathcal{Y}$, the dense mapping is the identity map $\pi(\mathcal{X}_i)=\mathcal{Y}_i$ .
We use this observation to define $\mathcal{L}_h$, the hard mapping loss:
\begin{equation}
\mathcal{L}_h = \frac{1}{|\mathcal{X}|}\sum_{i=0}^{|\mathcal{X}|-1}(1-\softP_{ii})
\end{equation}
which strives to have the cosine similarity between $x_i$,$y_i$ equal 1, de facto teaching the network to be invariant to the augmentation presented.

Furthermore, DWC presents a novel contrastive loss that motivates the embedding of a source point $x_i$ to be close to the neighbors of the corresponding target point $y_i$, while far from the points not in $y_i$'s proximity.
The neighborhood $\mathcal{N}$ of $y_i$ is defined to be its $k$ Euclidian nearest neighbors.
Defined by $A$ is the adjacency matrix of the target shape induced by $\mathcal{N}$ such that:
\begin{equation}
	A_{ij} = 1 \Leftrightarrow y_j \in \mathcal{N}_i
\end{equation}
We define $\overline{A}$ to be the negation of $A$, where $\overline{A}_{ij}=1$  for non-neighbors and $\overline{A}_{ij}=0$ otherwise.
Using the above notation, the positive $\mathcal{L}_{p}$ and negative $\mathcal{L}_{n} $ contrastive losses per-point are:
\begin{align}
\mathcal{L}_{p_{i}}=\sum_{\{j| A_{ij}=1\}}\max (0,m_p-\softP_{ij})\\
\mathcal{L}_{n_{i}}=\sum_{\{j| \overline{A}_{ij}=1\}}\max(0,\softP_{ij}-m_n)
\end{align}

Where $m_p$ and $m_n$ are the positive and negative margins.
The contrastive loss incorporates two significant advantages over previous methods:
\begin{itemize}
\item By applying the contrastive loss we fuse the topology of the graphs not only to the learning process but to the objective function itself.
\item As all target points takes part in $\mathcal{L}_{n}$, the network is exposed to all points, and not only to a chosen subset.
\end{itemize}
We further define $w_{Q_i}$ as the number of times $\mathcal{X}_i$ was chosen by $Q$

\begin{equation}
    w_{Q_i}=\sum_{j=1}^{|Q|}[Q_j=\mathcal{X}_i]
\end{equation} and factor the contrastive loss of $\mathcal{X}_i$ accordingly
\begin{align}
\mathcal{L}_{pQ}=\frac{1}{|Q|\sum\limits_{i,j} A_{ij}} \sum_{i}^{|\mathcal{X}|}w_{Q_i}\mathcal{L}_{p_{i}}\\
\mathcal{L}_{nQ}=\frac{1}{|Q|\sum\limits_{i,j} \overline{A}_{ij}} \sum_{i}^{|\mathcal{X}|}w_{Q_i}\mathcal{L}_{n_{i}}
\end{align}
By integrating $w_{Q_i}$ into $\mathcal{L}_{[n,p]}$ the contrastive loss is weighted based on the confidence of each source point. $\mathcal{L}_{[n,p] Q}$ compels DWC to define points as confident only if they are similar to the neighborhood of their matched target point, increasing the chances they are good "anchor" points for the alignment solution.

The final optimization function of DWC is:
\begin{equation}
 \mathcal{L}_C = \mathcal{L}_{h} + \mathcal{L}_{pQ} + \mathcal{L}_{nQ}
\end{equation}

\section{Experimentation}\label{section:Experimentation}

\begin{table*}[t!]
\centering
\resizebox{1.\textwidth}{!}{%
\begin{tabular}{l@{~}c@{~}c@{~}c@{~}c@{~}cc@{~}c@{~}c@{~}c@{~}cc@{~}c@{~}c@{~}c@{~}}
\toprule
& \multicolumn{4}{c}{Unseen point clouds}&& \multicolumn{4}{c}{Unseen categories}&& \multicolumn{4}{c}{Gaussian noise}\\
\cmidrule(lr){2-5}
\cmidrule(lr){7-10}
\cmidrule(lr){12-15}
\textbf{Model}  &\textbf{RMSE($R$)}&\textbf{MAE($R$)}&\textbf{RMSE($t$)}&\textbf{MAE($t$)}
&&\textbf{RMSE($R$)}&\textbf{MAE($R$)}&\textbf{RMSE($t$)}&\textbf{MAE($t$)}
&&\textbf{RMSE($R$)}&\textbf{MAE($R$)}&\textbf{RMSE($t$)}&\textbf{MAE($t$)}
\\
\midrule
ICP \cite{icp}
&16.62&14.53&0.710&0.628
&&17.13&16.25&0.892&0.813
&&11.23&10.45&0.101&0.095\\
 
Go-ICP \cite{icpl1}
&11.12&9.38&0.597&0.528
&&12.87&11.73&0.672&0.523
&&8.28&8.09&0.091&0.079\\

\noalign{\vskip 1.5mm}  
  \bottomrule 
  \noalign{\vskip 1.5mm} 
 
DCP \cite{dcp}
&6.12&5.83&0.024&0.015
&&7.73&6.12&0.037&0.025
&&6.52&5.34&0.034&0.031\\
 
PRNet \cite{prnet}
&4.24&3.95&0.012&0.011
&&5.10&4.11&0.020&0.017
&&7.14&6.35&0.021&0.019\\

PointNetLK \cite{pointnetlk}
&6.73&6.21&0.035&0.028
&&7.25&7.18&0.051&0.047
&&8.25&7.98&0.029&0.020\\
 
IT-Net \cite{itnet}
&5.43&5.11&0.063&0.051
&&6.12&4.98&0.086&0.081
&&8.63&7.72&0.035&0.031\\

RPM-net \cite{robustpointcloud}
&1.97&1.93&0.219&0.201
&&2.12&2.01&0.293&0.281
&&4.15&3.98&0.309&0.286\\

\noalign{\vskip 1.5mm}  
  \bottomrule 
  \noalign{\vskip 1.5mm} 
  
DWC (ours)
&\textbf{1.83}&\textbf{1.51}&\textbf{0.012}&\textbf{0.009}
&&\textbf{2.01}&\textbf{1.97}&\textbf{0.019}&\textbf{0.011}
&&\textbf{3.73}&\textbf{3.42}&\textbf{0.020}&\textbf{0.018}\\

\bottomrule
\end{tabular}%
}
\caption{ModelNet40 evaluations.
\hide{\textbf{Left}: train/test split of the entire dataset, 20\% of the shapes are not seen during training and being evaluated on. \textbf{Middle}: Category holdout, we train on a subset of ModelNet categories like airplanes and cars, but test on shapes with different geometric attributes like sinks and tables. \textbf{Right}: We add a random Gaussian noise from $\mathcal{N}(0,0.1)$ per point, and test with the split presented in the first experiment. }
DWC provides considerable performance gain in all evaluations both for $R$ and $t$. RPM-net is the only listed network that requires additional per-point normal to the surface information. MAE and RMSE stand for Mean Average Error and Root Mean Square Error respectively, lower is better.
}
\label{Tab:modelnetallresults}
\end{table*}

\subsection{Datasets and Metrics}
We compare DWC against two axiomatic methods, ICP \cite{icp} and Go-ICP \cite{icpl1}, and five different learnable models, DCP \cite{dcp}, PR-net \cite{prnet}, PointNetLK \cite{pointnetlk}, IT-Net \cite{itnet} and RPM-net \cite{robustpointcloud}. We used the publicly available code released by the authors for all of the above methods. \vspace{5.5pt}\\
The two datasets used for the evaluation are:
\paragraph{ModelNet40 \cite{modelnet}} a point cloud dataset contains 12,311 synthetically generated shapes from various categories. 

\paragraph{FAUST scans  \cite{faust}} Real human scans of 10 different subjects in 30 different poses each with about 80,000 points per shape. FAUST scans suffers from partiality, noise, and uneven sampling.\vspace{-7.0pt}\\

While ModelNet40 is extremely popular in this domain, we believe evaluating on non-synthetic datasets as FAUST is extremely important for the robustness assessment of the methods and the methods' ability to perform well in real-world scenarios.

Furthermore, unlike ModelNet40, which also has ground-truth information of the per-point normal to the surface, non-synthetic point-clouds can only approximate such information which is noisy and partial \cite{normal1,normal2,normal3}. 
In order to suite real-world datasets, our method uses only the Euclidean coordinates as inputs,
unlike RPM-Net \cite{robustpointcloud}, which also demands the existence of the point normal to the surface.

To assess the performance of DWC on the rigid alignment problem, we compare the predicted $R,t$ with the ground truth transformations using the root mean squared error (RMSE), and mean absolute error (MAE) metrics. This results in four metrics (two for each), where lower is better and zero is the optimal score. We measure the rotation difference using Euler angels and report the score in units of degrees.

\subsection{ModelNet40}\label{subsec:modelnet}

\begin{figure*}[ht]
\RawFloats
\begin{minipage}[b]{0.45\linewidth}
\centering
\includegraphics[width=\textwidth]{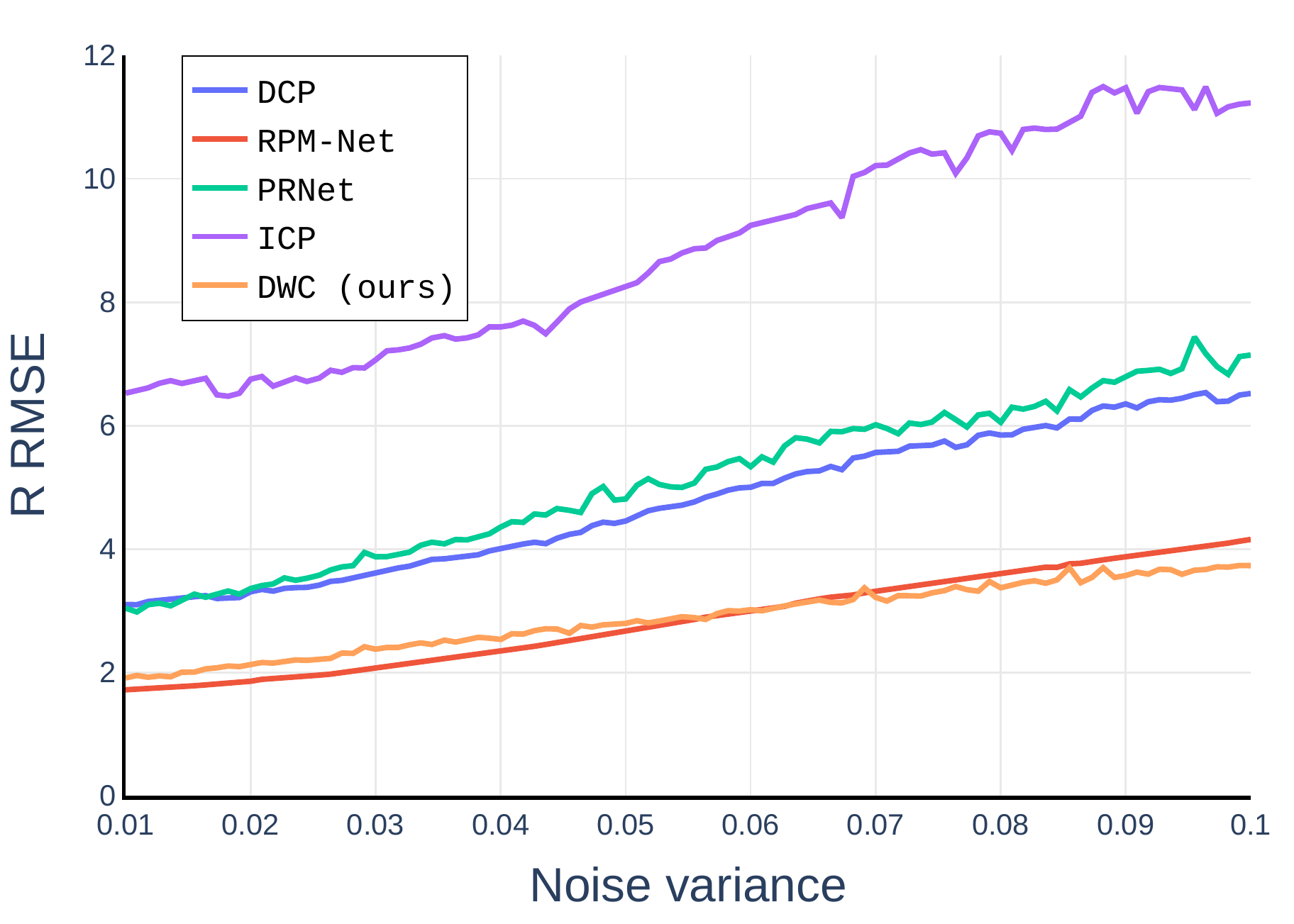}
\caption{Noise resilience evaluation.  The x-axis is the standard deviation of the Gaussian noise added per point $\mathcal{N}(0,\sigma)$, the y-axis is the angle of rotation error in RMSE.}
\label{fig:noise_resiliance}
\end{minipage}
\hspace{1.3cm}
\begin{minipage}[b]{0.45\linewidth}
\centering
\includegraphics[width=\textwidth]{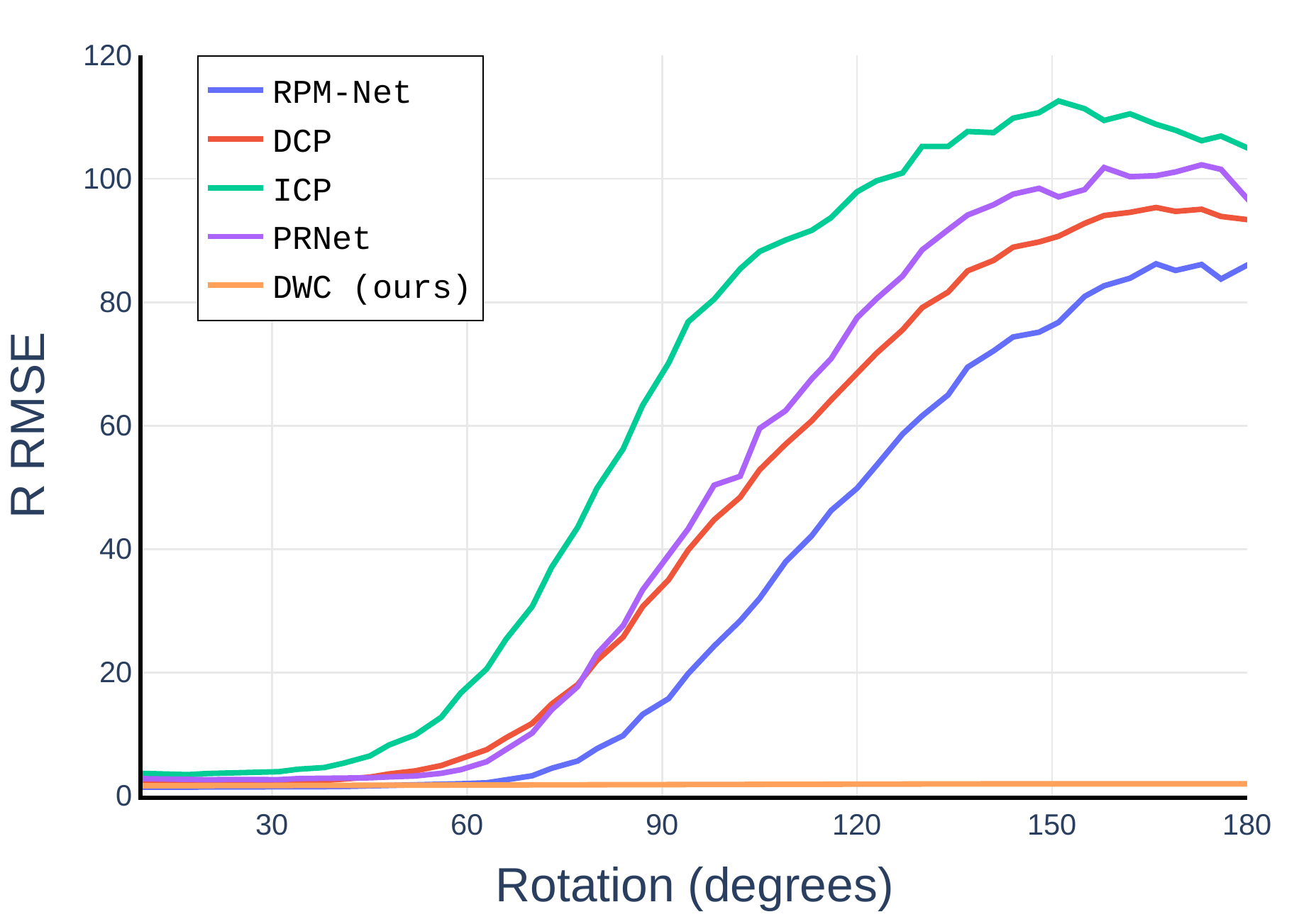}
\caption{Rotation resilience evaluation. The x-axis is  the  range of rotation $[0,R]$. The y-axis is the angle of   rotation  error  in RMSE. DWC is the only rotation invariant method.}
\label{fig:rot_resiliance}
\end{minipage}
\end{figure*}

ModelNet40 \cite{modelnet} consists of 12,311 CAD models from various categories, such as planes, furniture, cars, etc.  We do not use any input features other than the euclidian coordinates and unlike previous works \cite{pointnet,dcp,prnet}, we do not rescale the shapes to fit the unit sphere, as we found our network is resilient to the small scale-deformations this dataset holds.
To further emphasize the rotation invariance of DWC we provide figure \ref{fig:rot_resiliance}, where we evaluate the methods when applying rotations on the full spectrum of SO(3), that is, the random rotations are in $[-180,180]$ in contrast to \cite{pointnet,dcp,prnet,robustpointcloud}, where the rotations are limited to $[0,45-60]$. While testing with such a partial spectrum of SO(3) is an interesting proof of concept, most real-world applications encounter more extreme rotations, as in the case of multi-view registration for example.
In order to fairly compare the performance of DWC and the other methods, we compare the registration performance under rotations in $[0,60]$ for table \ref{Tab:modelnetallresults}, while showing the incompetence of all other architectures under large rotations in figure \ref{fig:rot_resiliance}.
During training we follow others \cite{pointnet,dcp,prnet} and subsample 1024 points randomly for each source shape $\mathcal{X}$. We then draw $R\in$ SO(3), $t\in \mathbb{R}^3$  (the deformation parameters), and set $\mathcal{Y}$ to be $T_{R,t}(\mathcal{X})$, the result of transforming $\mathcal{X}$ by $R,t$. For fair and full comparison, we re-train the baselines for each experiment until convergence with rotations in the relevant spectrum.

ModelNet40 experimentation is divided into 4 sub-experiments as follows:
\paragraph{Random train/test split} In this experiment the dataset is split randomly into 9843 train samples (80\% of the dataset), and taking the rest as test samples, samples that the network does not see in any deformation during training. 
The left part of \ref{Tab:modelnetallresults} presents the results on the described setting, where we present large margins in comparison to all other methods.
Specifically, we see an relative improvement of $7\%$ in RMSE($R$) compared to the second-best model (RPM-net).   

\paragraph{Unseen categories} To evaluate the generalization ability of DWC to scenarios where the test data statistics are not present during training, we split ModelNet40 such that some of the categories are not present during training. Specifically, we train on the first 20 categories and test on the rest. In the resulting split we train on categories such as planes, cars, but evaluate on chairs and plants, for example.
In the middle of table \ref{Tab:modelnetallresults} are the results for this experiment, where again, we present superior results compared to all others.
Besides DWC and RPM-net, all methods show great drops in performance compared to the previous experiment, varying from 20\% to 30\% relative degradation. We attribute our stability to the consensus voting, where we suggest-and-vote for multiple transformations, instead of solving one globally optimal alignment problem.

\paragraph{Random noise}
One of the biggest caveats of ICP is its lack of noise resilience, which may harm dramatically its results.
It is thus important to test noise durability.
We evaluate this by adding random Gaussian noise to the target points. In our setup, the added noise is 5 times higher than in previous papers to emphasize our noise resistance capabilities. In detail, we add to each point in the cloud a Gaussian noise sampled from $\mathcal{N}(0,0.01)$, and do not perform any noise clipping, unlike others. Without clipping, there may be significant outliers, as usually happens with real-world sensors. The right part of table \ref{Tab:modelnetallresults} shows the results of this test.
Like the two previous experiments, we present results that are up to 50\% better than previous state-of-the-art methods, with a substantial margin of 11\% to the second-best model. This advantage is due to the confidence sampling (Section \ref{subsec:softcorr}), which assigns low confidence to outliers, thus these points are not taken into account at the transformation computation. 

\paragraph{Noise and Rotation resilience}
To stress the most prominent traits of DWC, which are the rotation invariance and noise resilience, we provide figures  \ref{fig:noise_resiliance},\ref{fig:rot_resiliance}, comparing the effect of increasing the noise and rotation on the different methods.
From the noise resilience test, we see an almost linear dependence between the noise and $R$ RMSE for all methods, where RPM-Net shows better results for lower noise (up to $\sigma=0.532$) and DWC perform best from that point onward.\\
For the rotation experiment, each method was trained for each rotation factor until convergence, resulting in 100 experiments per method in the range of $[0,360]$. DWC shows 0.3\% degradation in the results, from 1.51 $R$ RMSE to 1.55 $R$ RMSE, while all others become irrelevant, with $R$ RMSE ranging from 80 to 120 when exposed to the full spectrum of SO(3). Such high $R$ RMSE reveals the impracticality of these methods under big deformations, as seen in figure \ref{fig:Tizer_pairs}.

\begin{table}[t!]
\centering
\resizebox{0.8\textwidth}{!}{%
\begin{tabular}{lwc{1.2cm}wc{1.2cm}p{0.1cm}wc{1.2cm}wc{1.2cm}}
\toprule
& \multicolumn{2}{c}{SO(3)\textsubscript{60}}&& \multicolumn{2}{c}{SO(3)}\\
\cmidrule(lr){2-3}
\cmidrule(lr){5-6}
\textbf{Model}  &\textbf{RMSE($R$)\ \ }&\textbf{RMSE($t$)}
&&\textbf{RMSE($R$)\ }&\textbf{RMSE($t$)}

\\

\midrule

ICP
&17.13&0.951&&95.36&1.232\\
 
Go-ICP
&16.52&0.837&&88.19&1.097\\

\noalign{\vskip 2mm}  
  \bottomrule 
  \noalign{\vskip 2mm} 
 
DCP
&8.37&0.091&&105.88&0.937\\
 
PRNet
&7.95&0.088&&83.12&0.957\\

PointNetLK
&6.12&0.097&&91.67&0.813\\
 
IT-Net
&4.32&0.155&&102.34&1.057\\

RPM-net
&15.79&1.175&&109.73&1.294\\

\noalign{\vskip 2mm}  
  \bottomrule 
  \noalign{\vskip 2mm} 
  
DWC (ours)
&\textbf{3.12}&\textbf{0.082}&&\textbf{3.29}&\textbf{0.091}\\

\bottomrule
\end{tabular}%
}
\caption{FAUST scans evaluation. When testing on the full spectrum of SO(3), all other methods fail to provide meaningful results. $SO(3)_{60}$ refers to rotations bounded to $[0,60]$.
}
\label{Tab:faustso3}
\end{table}

\subsection{FUAST scans}
\begin{figure}
\centering

\setlength\tabcolsep{0pt}

\begin{tabular*}{\textwidth}{
  @{\extracolsep{\fill}}
  >{\footnotesize}{l} 
  cccccc
}
&\multicolumn{2}{c}{ModelNet40 \cite{modelnet}}&\multicolumn{2}{c}{FAUST scans \cite{faust}}\\\\
Input
       &\includegraphics[width=0.25\textwidth,valign=c]{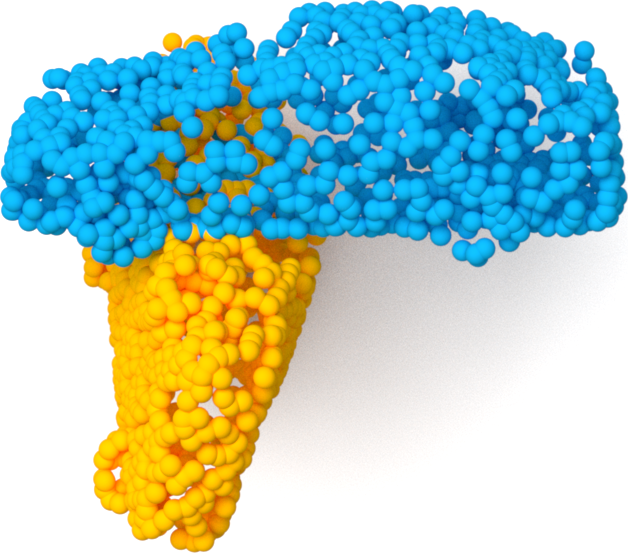}
       &\includegraphics[width=0.17\textwidth,valign=c]{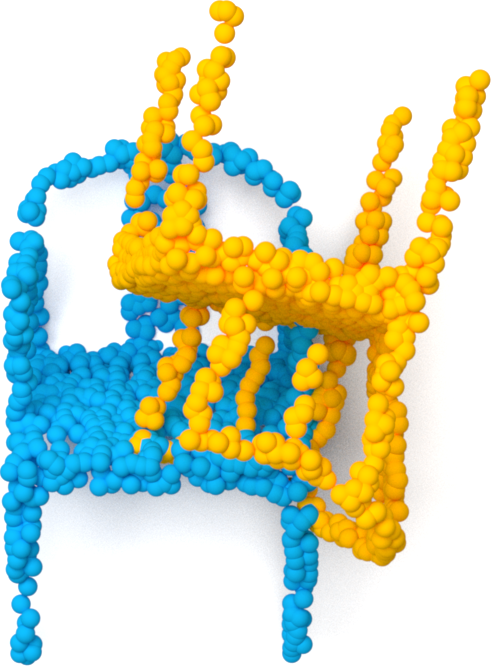} 
       &\includegraphics[width=0.15\textwidth,valign=c]{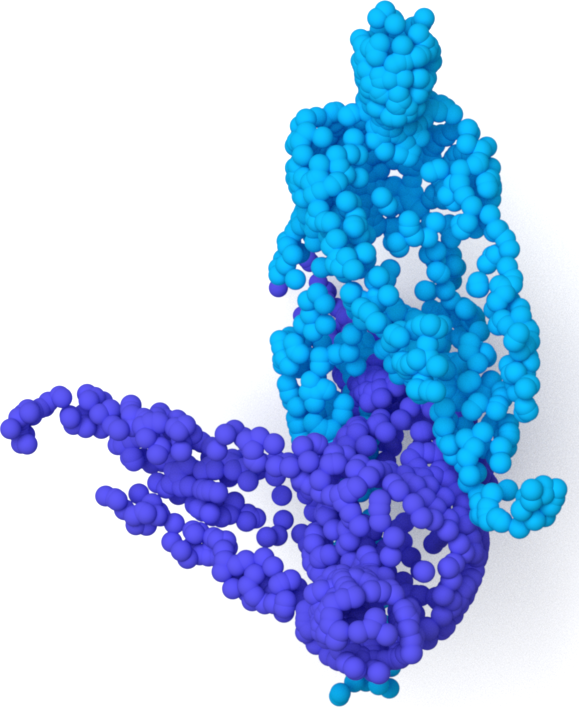}
       &\includegraphics[width=0.14\textwidth,valign=c]{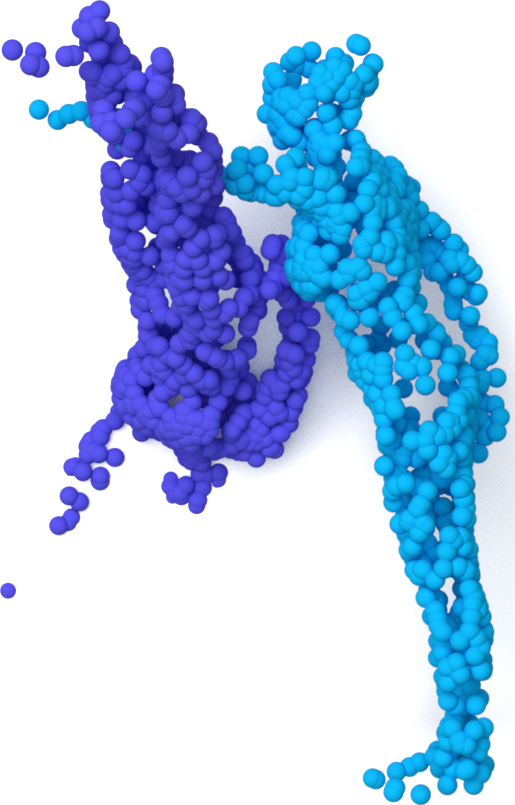}

\\ \addlinespace
Output
       &\includegraphics[width=0.28\textwidth,valign=c]{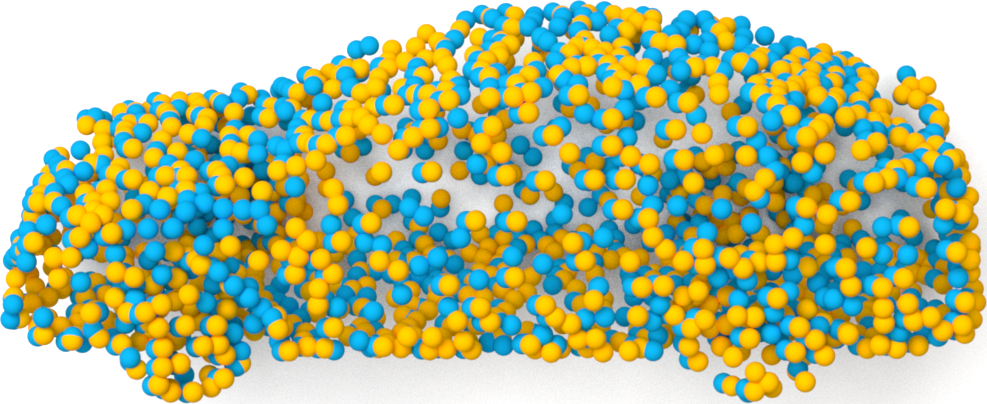}
       &\includegraphics[width=0.15\textwidth,valign=c]{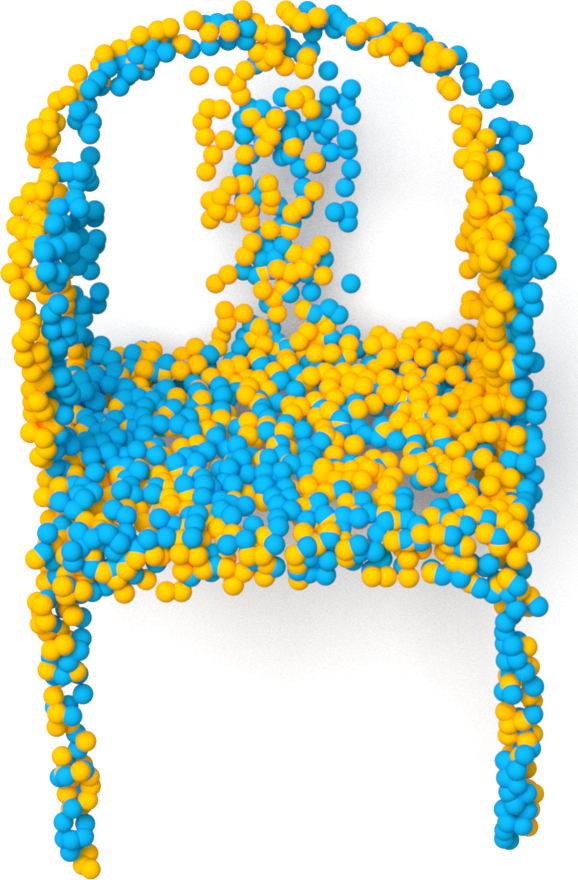} 
       &\includegraphics[width=0.09\textwidth,valign=c]{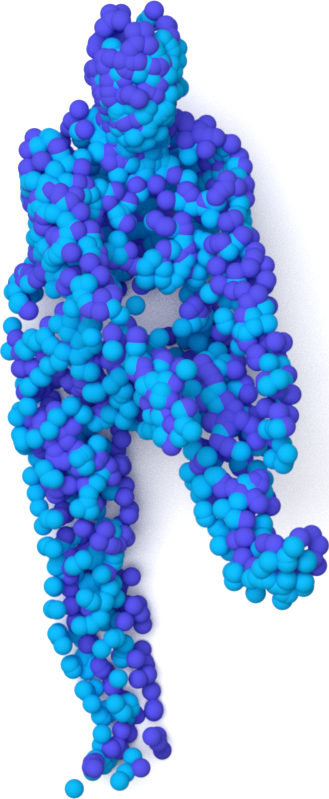}
       &\includegraphics[width=0.14\textwidth,valign=c]{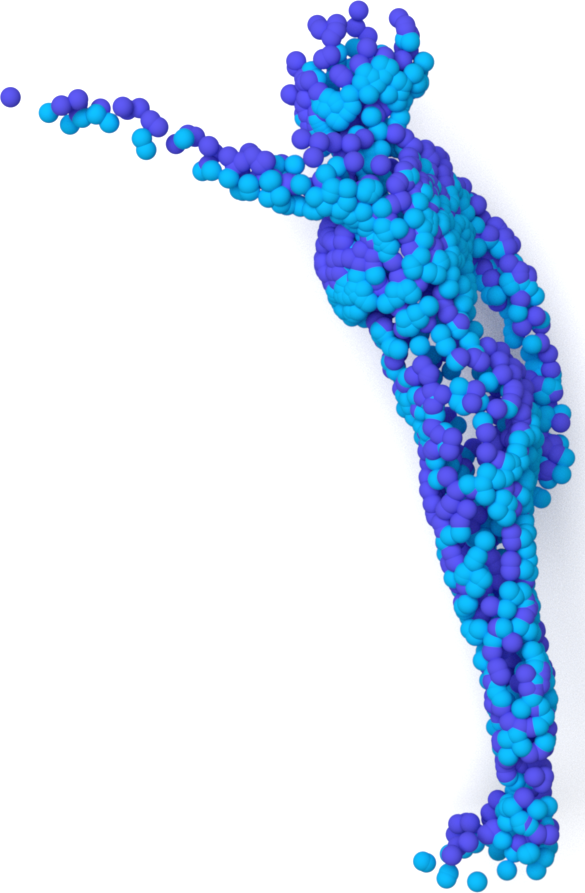}
\\
&\makecell{\ \ \ \quad Unseen \\ \quad \ \quad shape} & \makecell{\ \ Unseen \\ \ \  category} & \multicolumn{2}{c}{\makecell{\ \ \quad \ \quad Unseen \\ \ \quad \ \ \ \quad  dataset}}\\

\end{tabular*}
\caption{ \textbf{Generalization capability}. DWC alignment performance under three generalization use-cases: unseen shape from the trained category ("Unseen shape"), shape from an unseen category characterized by different statistical and geometric attributes ("Unseen category"), and shapes from a completely different dataset ("Unseen dataset").} \label{fig:generalization test}
\end{figure}

The FAUST dataset \cite{faust} contains 300 real human scans, without any post-acquisition filters, as denoising filters or shape completion. The resulting dataset contains noisy point-clouds with partial information in places that were occluded to the acquiring sensor. Evaluating rigid-alignment in real-world applications is crucial, thus, proving the ability to generate reliable results for a datasets such as FAUST is important.
In order to further examine the generalization ability of the evaluated models, we do not train on FAUST scans, and test the converged models trained on ModelNet40 (Figure \ref{fig:generalization test}).
In table \ref{Tab:faustso3} we present two evaluations, the first, where we test the models trained on rotations bounded to $[0,60]$ , and the second, where we evaluate the architectures after training on ModelNet40 on the full spectrum of SO(3).\\
RPM-net requires point-normals as input features. As these are not provided by FAUST, we use RNe \cite{normal1}, an axiomatic robust normal estimation for point clouds to make the inference on FAUST relevant for RPM-net also.
DWC presents superior results to all methods, with trends that correlate with ModelNet40 evaluations. As previously shown in figure \ref{fig:rot_resiliance}, all other methods fail when exposed to the full spectrum of SO(3). As to RPM-net, which was the second-best model on ModelNet40, we see a dramatic shift, emphasizing the disadvantage of normal-based methods in the point-cloud domain.

\subsection{Ablation}\label{subsec:ablation}
Table \ref{Tab:ablation} presents an ablation study for DWC, evaluated on the ModelNet40 train-test split, and FAUST scans datasets with rotations from the full spectrum of SO(3).

The following variants are considered: \vspace*{3pt}\\
(i) $RI$ features - We replace the $RI$ features\footnote{$RI$ features are not extra parameters but rather derived directly from the euclidean coordinates.} with the Cartesian coordinates of the shapes as the input descriptors to the $FEN$ module.\vspace*{3pt}\\
(ii) $DGCNN_{glob}$ - Discard the global feature vector generated by the $DGCNN_{glob}$ module. As a consequence, $GLF$ is discarded as well.\vspace*{3pt}\\
(iii) Static-Graph - We use the suggested Dynamic graph topology from DGCNN \cite{dgcnn} in the FEN instead the topology defined by the local Euclidian neighborhood.\vspace*{3pt}\\
(iv) $\mathcal{P}_m$ sampling - In order to evaluate the significance of the confidence sampling defined by $\mathcal{P}_m$, we replace $S$ with a uniform sampling over all source points.\vspace*{3pt}\\
(v) Consensus voting (full) - We replicate the inference procedure presented in DCP \cite{dcp}, and replace our consensus voting for $R,t$ with solving the $SVD$ on the entire points in the cloud.\vspace*{3pt}\\
(vi) Consensus voting ($top\textnormal{-}k$) - We replicate the inference procedure presented in PRNet \cite{prnet}, and replace our consensus voting for $R,t$ with solving the $SVD$ on the $top\textnormal{-}k$ source points with respect to the confidence metric\vspace*{3pt}\\
(vii) Contrastive loss - Our novel contrastive loss changes the training paradigm in the rigid correspondence domain to training directly the dense mapping. We test the effectiveness of this concept by replacing the contrasitve loss with $\mathcal{L}_2$ loss on $R,t$, as previous works have done:
\begin{equation}
    \mathcal{L} = ||R^T_{\mathcal{X}\mathcal{Y}}R^{gt}_{\mathcal{X}\mathcal{Y}} - I||^2 + ||t_{\mathcal{X}\mathcal{Y}} - t^{gt}_{\mathcal{X}\mathcal{Y}}||^2 
\end{equation}

The ablation study affirms the significance of all mentioned parts of the network. The use of RI features present dramatic performance enhancements, improving $R$ RMSE by 8 points of average. Nevertheless, we see that even without them, our results on the full spectrum are dramatically better than other methods (Figure  \ref{fig:rot_resiliance}). After the RI features, the biggest contributions come from the contrastive loss, consensus voting, and our confidence metric, all offer a paradigm shift to the mainstream approach of solving the rigid alignment problem. In the supplementary we provide a performance analysis of the baseline models when the input features are replaced with our $RI$ features.

\begin{table}[t!]
\centering
\resizebox{1\textwidth}{!}{%
\begin{tabular}{l@{\hspace{0.1em}}lwc{1.2cm}wc{1.2cm}p{0.1cm}wc{1.2cm}wc{1.2cm}}
\toprule
&& \multicolumn{2}{c}{ModelNet40}&& \multicolumn{2}{c}{FAUST}\\
\cmidrule(lr){3-4}
\cmidrule(lr){6-7}
&\textbf{Mode}  &\textbf{RMSE($R$)}&\textbf{RMSE($t$)}
&&
\textbf{RMSE($R$)}&\textbf{RMSE($t$)}

\\
\midrule
(i)&$RI$ features
&9.93&0.211&&11.07&0.309\\
 
(ii)&$DGCNN_{cls}$
&4.52&0.107&&5.12&0.103\\

(iii)&Static-Graph
&4.11&0.153&&5.72&0.160\\
 
(iv)&$\mathcal{P}_m$ sampling
&5.62&0.094&&6.92&0.097\\
 
(v)&Consensus (full)
&8.72&0.192&&9.87&0.199\\

(vi)&Consensus ($top\textnormal{-}k$)
&7.95&0.188&&8.34&0.157\\

(vii)&Contrastive loss
&9.45&0.197&&10.37&0.153\\

&Full method
&\textbf{1.83}&\textbf{0.012}&&\textbf{3.12}&\textbf{0.082}\\
 
\bottomrule
\end{tabular}%
}

\caption{Ablation study performed on ModelNet40 and FAUST scans. The "Mode" column states the switched \textbf{off} component.
}
\label{Tab:ablation}
\end{table}


\section{Summary}
\label{sec:Summary and Limitations}

DWC offers a paradigm shift for solving the rigid alignment problem by explicitly optimizing the dense map. In addition, unlike previous methods that solved for the globally optimal parameters which tend to be extremely sensitive to outliers, DWC presents a differentiable reduction procedure based on the confidence of each source point in its correspondence map, and weight the alignment solution accordingly. We exemplify under multiple experiments how these novelties present a significant boost in performance in all popular benchmarks, and, for the first time, solve the alignment problem for extreme rigid transformations in the full spectrum of SO(3).

\newpage
\clearpage
\nocite{leakyrelu,layernorm,pytorch}
{\small
\bibliographystyle{ieee_fullname}
\bibliography{egbib}
}

\end{document}